\documentclass{article}

\PassOptionsToPackage{numbers,sort&compress}{natbib}

 \usepackage[preprint]{neurips_2026}

\usepackage[utf8]{inputenc} % allow utf-8 input
\usepackage[T1]{fontenc}    % use 8-bit T1 fonts
\usepackage{hyperref}       % hyperlinks
\usepackage{url}            % simple URL typesetting
\usepackage{booktabs}       % professional-quality tables
\usepackage{amsmath}
\usepackage{amsfonts}       % blackboard math symbols
\usepackage{nicefrac}       % compact symbols for 1/2, etc.
\usepackage{microtype}      % microtypography
\usepackage{xcolor}         % colors
\usepackage{graphicx}

\title{Prisma-World: Camera-Controllable Multi-Agent Video World Model}

\author{Huiqiang Sun$^{1\ast}$, 
Zhan Peng$^{1\ast}$, 
Size Wu$^{2}$, 
Kun Wang$^{3}$, 
Kang Liao$^{2}$, 
Dianyi Wang$^{4}$, \\
\textbf{
Xingyu Zeng$^{5}$, 
Sheng Jin$^{6}$, 
Yangguang Li$^{7}$, 
Zhiguo Cao$^{1}$, 
Ziwei Liu$^{2}$, 
Wei Li$^{2\dagger}$} \\
$^{1}$School of AIA, HUST \quad 
$^{2}$S-Lab, NTU \quad \\
$^{3}$SenseTime Research \quad
$^{4}$FDU \quad
$^{5}$SUAT \quad
$^{6}$HKU \quad
$^{7}$CUHK
}

\begin{document}

\maketitle

\begingroup
\renewcommand{\thefootnote}{\fnsymbol{footnote}}
\footnotetext[1]{Equal contribution. \quad $^\dagger$ Corresponding author.}
\endgroup

\begin{abstract}
Video world models have made rapid progress in generating controllable visual experiences, but most of them still simulate the world from a single observer. Extending such models to multiple agents raises a central challenge: if each agent's future state is generated independently, overlapping views may instantiate different versions of the same scene, leading to inconsistent objects, layouts, and appearances across agents. Conventional camera conditioning controls individual trajectories, but it does not explicitly couple the generation of views that should agree under shared scene geometry. We introduce \textbf{Prisma-World}, a camera-controllable multi-agent world model that formulates multi-agent generation as a joint geometry-aware denoising process for cross-view consistency. Prisma-World processes all agent videos within one full-attention sequence, uses a multi-agent RoPE design to distinguish agent identities while preserving synchronized temporal coordinates, and injects relative camera geometry into attention to bias overlapping viewpoints toward shared scene evidence. To further strengthen multi-view consistency and enhance global spatial perception, we augment our framework with an overlap-decaying curriculum training paradigm alongside minimap-conditioned structural guidance. To facilitate the training and evaluation of multi-agent models, we introduce \textbf{PrismaDataset}, a large-scale UE5 dataset with panoramic acquisition across diverse scenes, composable multi-agent view groups with flexible agent counts and complex camera trajectories, and precise camera/action annotations for consistency training and evaluation.  Experiments show that a single Prisma-World model can generate high-fidelity multi-agent videos with flexible agent numbers, camera controllability, improved cross-view consistency, and spatial grounding under minimap guidance. Project Page: \url{https://huiqiang-sun.github.io/prisma-world/}.
\end{abstract}    
\begin{figure*}
    \centering
    \includegraphics[width=\linewidth]{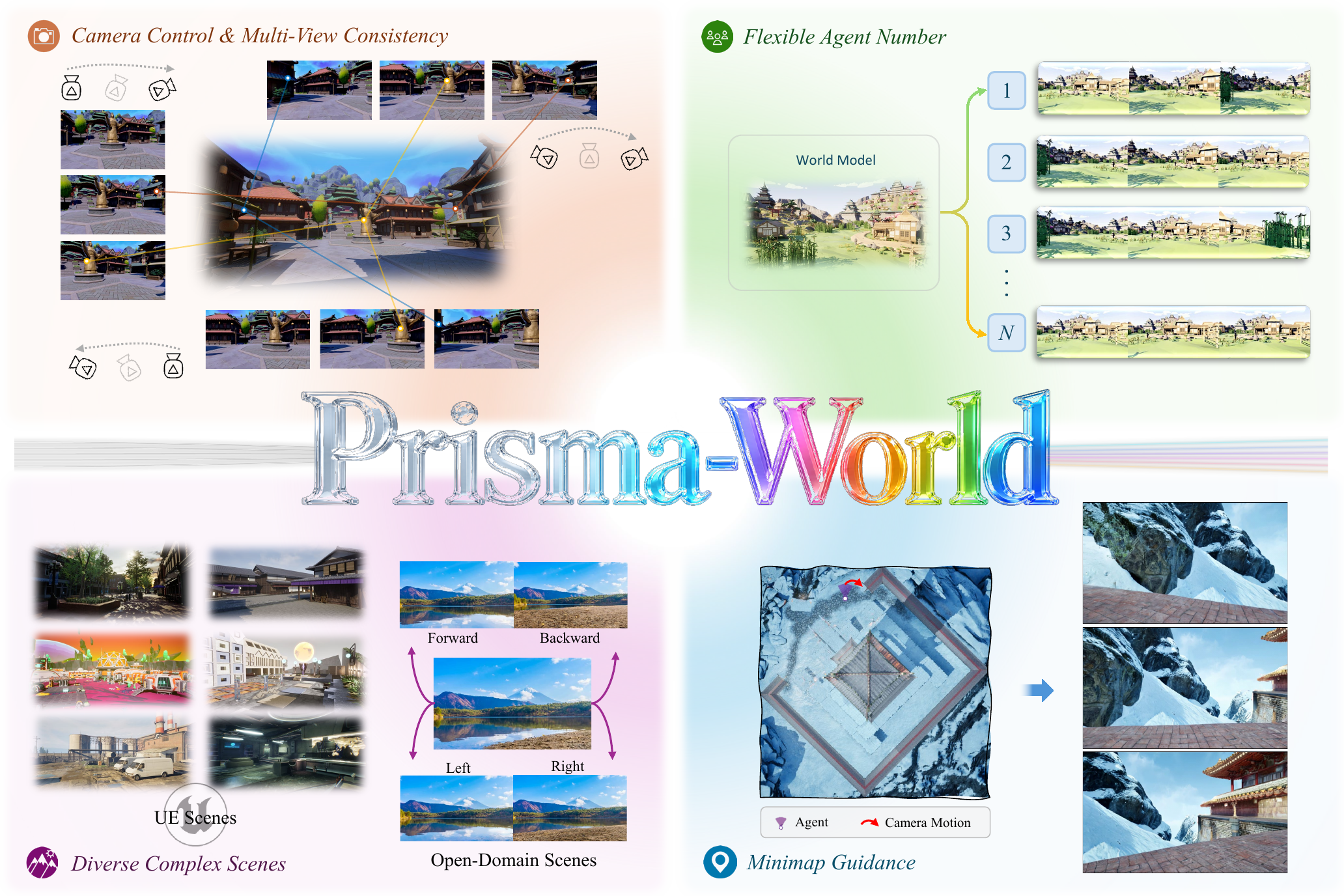}
    \caption{In this paper, we introduce \textbf{Prisma-World}, a camera-controlled multi-agent world model capable of synthesizing multi-agent videos within complex scenes while preserving multi-view consistency. Furthermore, our model offers the flexibility to specify the number of output agents and supports minimap conditioning to provide explicit local spatial structural guidance for each agent.}
    \label{fig:teaser}
    \vspace{-2mm}
\end{figure*}

\section{Introduction}
\label{sec:intro}

Generative video world models~\cite{kong2024hunyuanvideo, li2025hunyuan, tang2025hunyuan, mao2025yume, he2025matrix, wang2026matrix, decart2024oasis, genie3, lingbot-world} have recently achieved unprecedented breakthroughs in high-fidelity video synthesis. Trained on large-scale monocular videos with action or camera annotations, these models have demonstrated remarkable capabilities in video generation~\cite{sora, seedance2026seedance}, agent behavior control~\cite{magne2026nitrogen, cen2025worldvla, zhang2026dreamvla}, and camera trajectory manipulation~\cite{he2024cameractrl, Yu_2025_ICCV}. However, existing architectures are predominantly designed to simulate the world from the perspective of a single, isolated agent. In contrast, numerous real-world applications, such as advanced gaming environments and autonomous navigation systems, are inherently populated by multiple interacting entities. This reality necessitates models capable of simultaneously generating multi-agent videos within a shared environment. Evidently, the vast majority of current paradigms lack the fundamental capacity to synthesize multi-agent visual streams, let alone guarantee multi-view consistency across different agents. Consequently, the development of multi-agent video world models remains a largely unexplored frontier.

Multi-agent video world modeling exposes a structural limitation in existing generation paradigms. A direct extension of a single-agent model samples each agent independently, which implicitly factorizes the joint world into separate visual streams and gives the model no mechanism to enforce a shared scene state. As a result, two agents looking at the same building, road, or object may generate different appearances because each video is sampled as its own world instance. Another possible extension is to concatenate all videos into a longer sequence, but this introduces a different ambiguity: the model must know which tokens correspond to different agents, which frames are synchronized across agents, and which views should be consistent with camera geometry. Meanwhile, existing camera-conditioned video methods mainly use camera parameters to control a single trajectory; they do not explicitly use relative camera pose to regularize consistency between overlapping views. These issues make multi-agent generation a cross-view consistency problem rather than only a data or compute scaling problem.

In this paper, we propose \textbf{Prisma-World}, a camera-controllable multi-agent world model that treats all agents as coupled observations of one shared scene. Prisma-World performs joint denoising over all agent videos in one full-attention sequence, enabling cross-view consistency to emerge during generation. To make this joint sequence interpretable to the model, Prisma-World introduces a multi-agent RoPE design that separates agent identities in the positional space while keeping equal-time frames temporally aligned. To further align consistency with scene geometry, Prisma-World injects relative camera transformations into attention, encouraging nearby or overlapping views to rely on shared scene evidence while allowing distant views to preserve viewpoint-specific content. This consistency-oriented architecture is paired with an overlap-decaying curriculum that first learns from strongly overlapping views and then adapts to harder trajectory combinations to enhance the ability of multi-view consistency. Additionally, Prisma-World accommodates an optional global minimap input as a structural condition. This macroscopic prior provides explicit scene layout guidance, significantly augmenting the spatial perception and grounding during multi-agent generation.

To train and evaluate this setting, we construct \textbf{PrismaDataset}, a large-scale UE5-based dataset tailored for controllable multi-agent world modeling. PrismaDataset covers diverse complex scenes and provides RGB videos, precise camera poses, agent actions, and optional top-down minimaps. By recording panoramic videos and projecting them into synchronized perspective views, we can compose multi-agent video groups with flexible agent counts, complex camera trajectories, and controllable degrees of visual overlap. We also construct a benchmark, MultiAgentBench, tailored to quantitatively assess the multi-view scene consistency across different generated agents. Experiments on the proposed benchmark show that Prisma-World can generate high-quality multi-agent videos in complex environments, while flexibly scaling to varying numbers of agents using a single model. It consistently outperforms both single-agent and concurrent multi-agent baselines in terms of cross-view consistency, and further benefits from minimap guidance, which improves spatial grounding and scene-level coherence. In summary, our key contributions are multifold:

\begin{itemize}
    \item We propose Prisma-World, a multi-agent world model with multi-agent RoPE and camera-aware cross-view consistency modeling. It is capable of synthesizing diverse and complex scenes while guaranteeing multi-view consistency and offering the flexibility to dynamically specify the number of generated agents.
    \item We introduce training and conditioning designs for robust multi-agent generation, including an overlap-decaying curriculum for enhancing multi-view consistency and minimap-based spatial guidance.
    \item We build PrismaDataset, a large-scale panoramic UE5 dataset with precise camera/action annotations, flexible multi-agent view composition, complex camera trajectories, and a benchmark for evaluating multi-agent view consistency.
\end{itemize}
\section{Related Work}
\label{sec:related_work}

\noindent\textbf{Camera-Controlled Video Generation.}
Diffusion-based video generative models~\cite{blattmann2023stable, chen2024videocrafter2, hong2022cogvideo, wan2025} have been widely extended with camera conditions for T2V~\cite{he2024cameractrl, wang2024motionctrl}, I2V~\cite{Yu_2025_ICCV, gu2025diffusion}, or V2V~\cite{bai2025recammaster, van2024generative} models, enabling the synthesis of videos following specific camera trajectories. Early methods typically encoded camera parameters directly into features for injection into DiT modules~\cite{bai2025recammaster, wang2024motionctrl}, or converted them into perspective fields~\cite{puffin}, ray-maps~\cite{gao2024cat3d}, or Plücker coordinates~\cite{bahmani2025ac3d, he2024cameractrl, liang2025wonderland} based on camera origins and directions. To improve control precision and spatial perception, relative camera encoding techniques emerged~\cite{li2026cameras, miyato2024gta}. Specifically, PRoPE~\cite{li2026cameras} integrates the relative relationships between camera parameters across different frames into the attention mechanism, enhancing multi-view understanding and view synthesis. UCPE~\cite{zhang2025unified} further incorporates control signals for the intrinsic and distortion parameters. However, these methods remain restricted to single-agent scenarios and cannot support camera-controlled video generation involving multiple agents.

\noindent\textbf{Video World Model.}
Video World Models~\cite{lingbot-world, he2025matrix, li2025hunyuan, tang2025hunyuan, xiang2024pandora} aim to learn environment dynamics from visual sequences and conditions such as text, images, actions, or camera motion. Early world models~\cite{ha2018world, hafner2023mastering} primarily served model-based reinforcement learning, facilitating agent planning by learning latent state transitions. With the recent advancement of generative models, some works treat video generation models as implicit world simulators~\cite{sora, feng2026matrix, valevski2025diffusion, rigter2024avid, alonso2024diffusion}, emphasizing visual realism, temporal consistency, and physical plausibility. For instance, Sora~\cite{sora} has demonstrated the potential of video generation in modeling motion, scene transitions, and object interactions, while the Genie series~\cite{genie3} introduced interactive control, enabling models to learn actionable environmental dynamics from unlabeled videos. Subsequent works have further improved critical characteristics of world models—such as physical consistency~\cite{song2025physical, wang2025enhancing, yuan2026inference, zhu2025aether}, long-term memory~\cite{hong2025relic, wu2026infinite, yu2025context, chen2026out, chou2025captain, wu2026video}, and efficient autoregressive generation~\cite{chen2024diffusion, huang2026self, yin2025slow}—enabling diverse downstream applications in embodied AI~\cite{chi2025wow, ali2025world, li2026causal, agarwal2025cosmos}, autonomous driving~\cite{russell2025gaia}, and gaming~\cite{decart2024oasis, tang2025hunyuan, he2025matrix, valevski2025diffusion}. Despite these impressive advancements, most existing methods operate under a single-agent assumption, neglecting the generation and control of multiple agents within the same environment. To address this, we propose a camera-controlled multi-agent world model framework capable of generating multi-agent videos in complex environments while ensuring consistency across a variable number of agents and multi-view perspectives.

\noindent\textbf{Multi-Agent Video Models.}
Despite the rapid progress in video world models, multi-agent generation remains an under-explored frontier. Leveraging advancements in video generation, early works attempted to generate multi-view videos from fixed perspectives~\cite{bai2025syncammaster} or utilized camera parameters to synthesize novel views with simple trajectories starting from a single image~\cite{kuang2024collaborative, xu2024cavia}. However, these methods rely on rudimentary camera control—often restricted to static viewpoints or simple shifts—failing to achieve interactive control over multiple agents or maintain multi-view consistency during complex motions. IC-World~\cite{wu2025ic} employs reinforcement learning to enhance multi-view consistency, while it lacks explicit camera-based control, thereby sacrificing the interactivity essential for a world model. Furthermore, its reliance on multi-view concatenation limits its scalability to a larger number of agents. Similarly, MultiVerse~\cite{enigma2025multiverse} simulates a two-agent racing scenario but remains constrained by its limited agent count and lack of scene diversity.

Recently, several concurrent works have explored multi-agent world models. Solaris~\cite{savva2026solaris} introduced a two-agent interaction model within the Minecraft environment, while MultiWorld~\cite{wu2026multiworld} attempted to achieve flexible and scalable multi-agent capabilities by compressing multi-view information. However, these methods are confined to a single gaming scene (e.g., Minecraft or It Takes Two), failing to generalize to more complex environments. Furthermore, they cannot leverage camera parameters to enable multi-agent spatial awareness. In contrast, Prisma-World enables camera-controlled multi-agent video generation across complex scenes. By utilizing the relative information of camera parameters, Prisma-World ensures multi-view consistency across overlapping regions within a shared environment, while a single trained model can flexibly accommodate a variable number of agents. Additionally, we introduce PrismaDataset, a large-scale multi-view dataset characterized by rich scene categories and precisely annotated camera poses. It supports the construction of data groups with an arbitrary number of agents, providing a solid foundation for model training. 
\section{Prisma-World: Multi-Agent World Model}
This section provides a detailed overview of our proposed multi-agent model, Prisma-World. In Sec.~\ref{sec:method_1} and Sec.~\ref{sec:method_2}, we elaborate on the model architecture, specifically detailing the processing of multi-agent video inputs and the control of output videos via camera parameters. Subsequently, Sec.~\ref{sec:method_3} outlines the training strategies employed during training on the PrismaDataset to enhance multi-agent model performance. Finally, Sec.~\ref{sec:method_4} introduces the integration of global minimap information to guide the generative process of the model. 

\subsection{Model Framework}
\label{sec:method_1}
Prisma-World is designed based on a simple observation: multi-agent video generation is not equivalent to running a single-agent video model multiple times. If each agent is sampled independently, the model factorizes the joint distribution into isolated streams and has no mechanism to negotiate shared objects, repeated structures, or overlapping regions. Conversely, merely concatenating multiple videos into a longer sequence is also insufficient, because the model must know which tokens belong to different agents, which frames are temporally synchronized, and which views should communicate according to camera geometry. Prisma-World addresses these issues by turning multi-agent generation into a single geometry-aware denoising process. Given $N$ controllable agents, the model jointly denoises all agent frames with full attention, uses an agent-aware positional system to preserve temporal correspondence across agents, injects relative camera geometry into attention, and optionally conditions each agent on a local minimap prior.

The core modeling challenge is to represent multiple agents as different observations of one shared world rather than as independent videos. Let $\mathbf{x}^{n}=\{x^{n}_{1},\ldots,x^{n}_{T}\}$ denote the video observed by agent $n$, and let $\mathbf{c}^{n}$ denote its camera trajectory and control signals. A straightforward single-agent extension would model
\begin{equation}
    p(\mathbf{x}^{1:N}\mid \mathbf{c}^{1:N})
    \approx
    \prod_{n=1}^{N} p(\mathbf{x}^{n}\mid \mathbf{c}^{n}),
\end{equation}
which is easy to implement but removes the statistical dependence between views. Under this factorization, two agents observing the same region may independently sample different textures or object layouts. Prisma-World instead models the joint distribution by processing all agent tokens in one denoising network,
\begin{equation}
    p_{\theta}(\mathbf{x}^{1:N}\mid \mathbf{c}^{1:N})
    =
    p_{\theta}(\mathbf{x}^{1},\ldots,\mathbf{x}^{N}\mid \mathbf{c}^{1},\ldots,\mathbf{c}^{N}),
\end{equation}
so that cross-agent consistency can be formed inside the generative process.

\begin{figure*}[t]
    \centering
    \includegraphics[width=\linewidth]{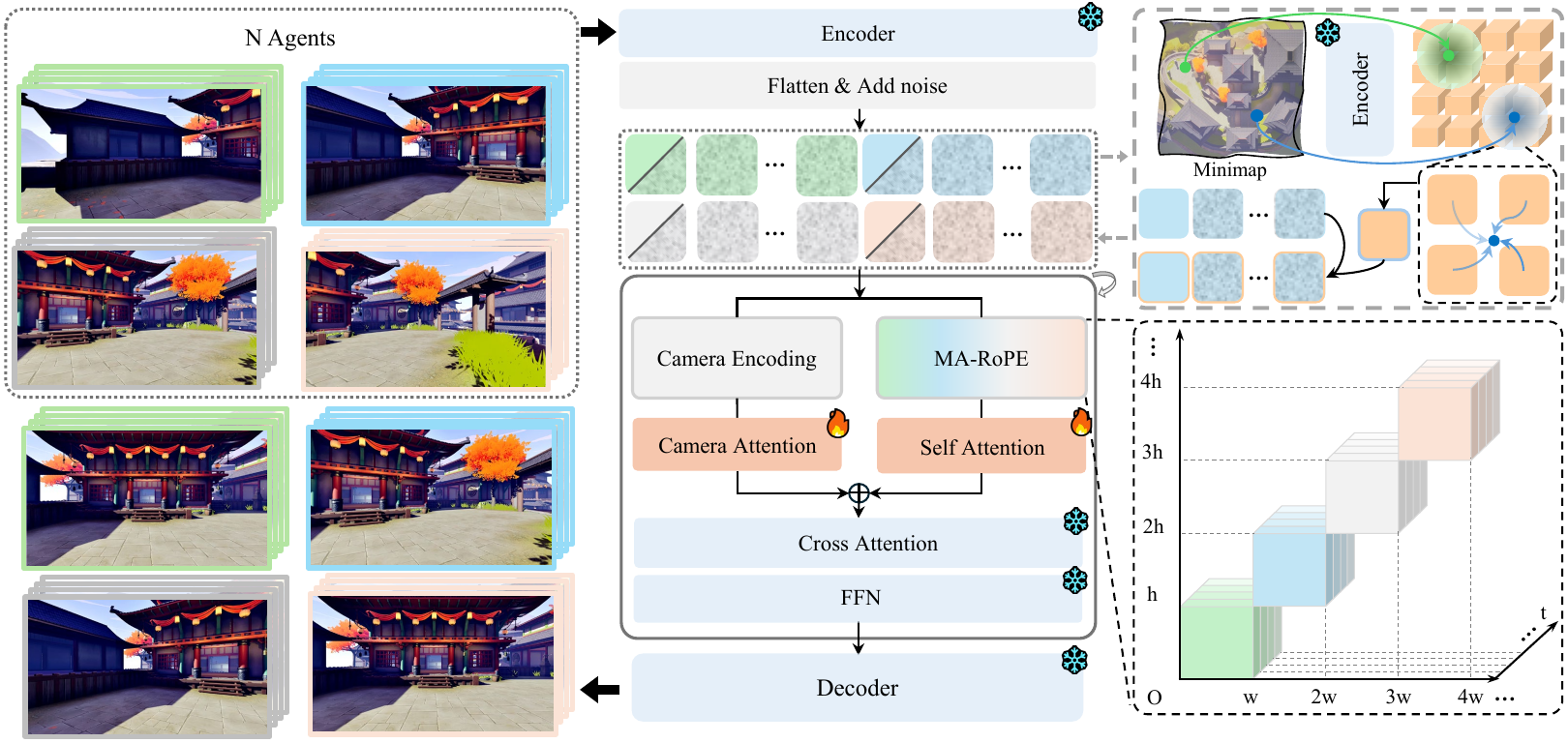}
    \caption{
        \textbf{Method overview.}~Prisma-World generates multi-agent videos within a single joint denoising process. Our MA-RoPE keeps frames at the same time step aligned across agents while distinguishing tokens from different agents. The minimap branch provides local spatial guidance by projecting each agent position onto a top-down minimap and injecting the extracted layout feature into the corresponding agent tokens.
    } 
    \label{fig:method}
    % \vspace{-1mm}
\end{figure*}

The overall model structure of Prisma-World is shown in Fig.~\ref{fig:method}. Specifically, Prisma-World builds on a pre-trained Wan2.2-TI2V-5B model~\cite{wan2025}. Given $N$ agent videos $\{V_i\}_{i=1}^N$, they are first encoded into latent representations $x$ via a frozen Wan VAE encoder. Subsequently, a patch embedding operation maps these latents to $z \in \mathbb{R}^{N \times C \times T \times H \times W}$. During the noising process, we maintain the first-frame tokens of each agent as clean, while adding noise to the remaining tokens. To simultaneously process the video content of all agents and establish inter-agent correlations, we flatten the latent features into a joint sequence of length $T \times N \times H \times W$. This joint token sequence is then passed into the subsequent DiT blocks for attention computation. Unlike independently denoising each agent, this design facilitates the joint generation of all agents within a single full-attention mechanism, allowing distinct viewpoints to directly exchange information throughout the denoising process. 

However, joint full attention only works if the token coordinates encode both agent identity and synchronized time. Standard 3D RoPE~\cite{su2024roformer} is designed for a single video and assigns coordinates according to time, height, and width. Directly reusing it for multi-agent videos introduces ambiguity: tokens from different agents at the same image location may receive identical spatial coordinates, while treating agents as extra temporal frames would destroy the fact that frame $t$ of different agents is synchronized. Therefore, we exploit a multi-agent RoPE (MA-RoPE) coordinate that separates agents in the spatial embedding while keeping their temporal coordinates aligned. Specifically, for a token indexed by time $t$, agent $n$, latent height $h$, and latent width $w$, we assign 
\begin{equation}
    \mathrm{RoPE}(t,n,h,w) = R_t(t) \oplus R_y(nH+h) \oplus R_x(nW+w) \, ,
\end{equation}
where $R_t$, $R_y$, and $R_x$ denote the rotary positional encodings along the temporal, height, and width dimensions, respectively. This design yields a critical property: when two agents are at the same latent time frame, they share identical temporal coordinates, resulting in a temporal relative offset of zero; meanwhile, since the spatial coordinates incorporate $nH$ and $nW$, different agents remain distinguishable within the RoPE space. In other words, Prisma-World structures the multi-agent videos into a joint spatial-temporal grid that is temporally aligned yet agent-distinguishable. Since all agent tokens reside within a single full-attention sequence, each viewpoint can attend not only to its own temporal context but also directly to the synchronous or adjacent tokens of other agents. This modeling paradigm ensures that cross-view consistency does not rely on post-processing; rather, it emerges naturally during the attention computation at each denoising step. 

\subsection{Camera-Aware Cross-View Consistency Modeling}
\label{sec:method_2}
To enable camera control over multi-agent videos, inspired by relative camera parameter encoding methods such as PRoPE~\cite{li2026cameras} and UCPE~\cite{zhang2025unified}, we construct a camera projection matrix $P$ based on the corresponding camera extrinsics. Concurrently, we integrate a parallel self-attention layer alongside the standard self-attention within the DiT module, explicitly incorporating camera coordinate transformations into the attention queries and keys. For a query derived from viewpoint $i$ and a key from viewpoint $j$, the camera-aware attention can be formulated as:
\begin{equation}
    \ell_{ij} = \frac{(P_i^{\top} q_i)^{\top} (P_j^{-1} k_j)}{\sqrt{d}} = \frac{q_i^{\top} (P_i P_j^{-1}) k_j}{\sqrt{d}} \, ,
\end{equation}
where $P_i P_j^{-1}$ denotes the relative camera transformation from viewpoint $j$ to viewpoint $i$. This formulation demonstrates that the attention weights are no longer determined solely by the similarity of appearance tokens; rather, they are concurrently modulated by the relative camera poses between the two viewpoints. When the camera positions and orientations of two agents are in proximity, their relative transformation $P_i P_j^{-1}$ approaches the identity transformation, prompting the attention mechanism to establish stronger correlations between these views. Conversely, when the viewpoints diverge significantly, the model is capable of expressing more pronounced perspective differences. In this manner, Prisma-World simultaneously models content similarity and camera geometry within the full-attention multi-agent token sequence, thereby encouraging spatially adjacent viewpoints to generate highly consistent scene content.

\subsection{Training Strategies}
\label{sec:method_3}
\noindent\textbf{Overlap-Decaying Curriculum Training.}
Training a multi-agent world model directly on arbitrary trajectories is unstable because useful consistency supervision is sparse. If two sampled agents rarely observe overlapping content, the model receives little signal about how to align their generations; if all views overlap strongly, the model may overfit to near-duplicate views and fail under diverse trajectories. Prisma-World therefore uses an overlap-decaying curriculum constructed from the PrismaDataset. Our dataset captures panoramic videos of the scenes and derives perspective videos with substantial visual overlap from these panoramas, which serve as training samples in the initial stages. During training, we progressively decrease the content overlap among distinct videos within the data samples, transitioning to samples that feature more complex and relatively independent camera trajectories. Compared to training directly on the non-overlap data, this curriculum learning strategy significantly facilitates the acquisition of multi-view consistency. 

\noindent\textbf{Variable Agent Training.}
We also train Prisma-World to support a variable number of agents. Since all agent tokens are processed in one full-attention sequence, the effective token length grows with $N$. A fixed noise schedule can therefore lead to different denoising difficulty across agent counts: small-$N$ batches may be over-regularized, while large-$N$ batches may become harder to stabilize. To reduce this mismatch, we use a dynamic noise-shift strategy that adjusts the flow-matching noise shift according to the current sequence length. As the number of agents increases and the token sequence lengthens, the noise schedule is increases linearly. This dynamically mitigates the training distribution shifts that occur across varying multi-agent configurations.

\subsection{Minimap Condition}
\label{sec:method_4}
Camera trajectories and first-frame conditions are often insufficient to specify the global scene layout around multiple agents. A first frame anchors only the currently visible region, and a text prompt usually describes the scene category rather than the precise spatial structure. This limitation becomes more severe in multi-agent generation: different agents may start from different locations, and the model must infer what should appear around each agent while maintaining one coherent scene. To this end, Prisma-World introduces an optional minimap condition to provide this spatial prior.

The minimap branch converts global layout information into agent-specific local guidance, as shown in Fig.~\ref{fig:method}. Given a minimap image $M$, we encode it with the VAE encoder to obtain a spatial feature map. For agent $n$, we project its initial world coordinate onto the minimap coordinate and extract a local feature $m_n$ using a Gaussian window centered at the projected point. The embedding $m_n$ is then injected into the tokens of agent $n$ before self-attention. In this way, the model receives a local layout hint for each agent while still generating all agents through the same joint attention process.

\section{Experiment}
\subsection{PrismaDataset}
We construct PrismaDataset in Unreal Engine 5 to provide multi-agent videos with accurate camera and action annotations. For each static scene, we use NavMesh in UE5 to identify navigable regions and deploy camera-equipped agents to automatically roam through the environment. During data collection, we record RGB videos, camera poses, agent actions, and optional top-down minimaps. This automated pipeline enables diverse trajectories across complex scenes while avoiding the cost and limited controllability of real-world multi-camera capture.

A key advantage of PrismaDataset is its panoramic acquisition strategy. We record panoramic videos and project them into five synchronized perspective views with different yaw offsets~\cite{veicht2024geocalib, puffin}, including the center view and $30$ and $60$ degrees to the left and right, denoted as $C$, $L_{30}$, $R_{30}$, $L_{60}$, and $R_{60}$. These views share the same underlying trajectory but contain different degrees of visual overlap, making them well-suited for training and evaluating multi-view consistency. Under this setting, PrismaDataset contains panoramic data from $96$ scenes and perspective data from $79$ scenes, totaling $620$ hours of video. More importantly, thanks to the panoramic video recording, our dataset can generate multi-agent video data groups of arbitrary sizes, featuring complex camera trajectories and precise camera annotations. This significantly elevates the overall diversity of the dataset. 

\begin{figure*}[t]
    \centering
    \includegraphics[width=\linewidth]{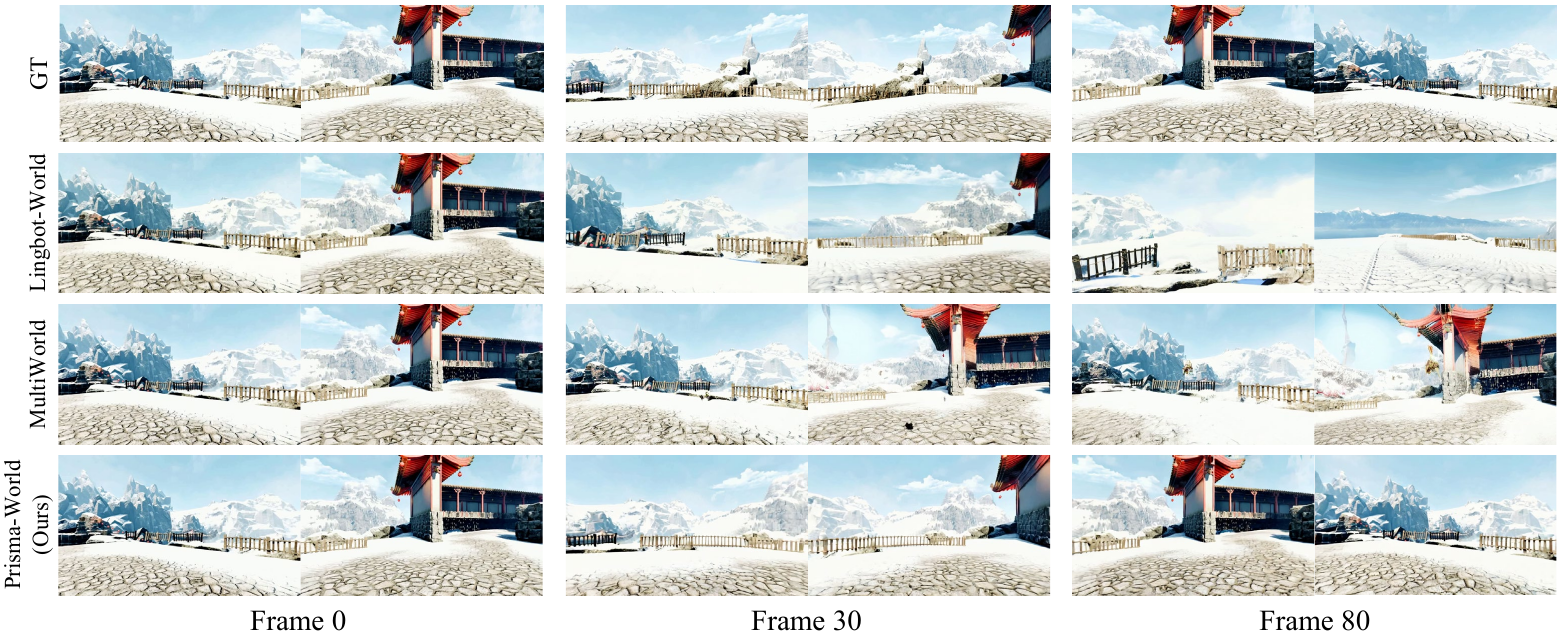}
    \caption{
        \textbf{Qualitative comparison.}~Compared to other baselines, our model can synthesize high-quality multi-agent videos with precise camera control while preserving multi-view consistency.
    } 
    \label{fig:compare}
\end{figure*}

\subsection{Experiment Settings}
\noindent\textbf{Evaluation Benchmark.}
We construct an evaluation benchmark, MultiAgentBench, using the panoramic images captured in the PrismaDataset. We designate a subset of the scenes from the PrismaDataset as the test set. Within this test data, for each acquired panoramic image, we construct test samples based on the smooth transitions across five predefined viewpoints ($C$, $L_{30}$, $L_{60}$, $R_{30}$, and $R_{60}$). The camera trajectories for each agent video can be arbitrarily composed and sequenced from these five distinct views. This experimental design enables the rigorous evaluation of multi-agent view consistency across various dimensions. Specifically, it assesses single-agent revisit consistency, synchronous multi-agent consistency when observing unseen regions, and asynchronous multi-agent consistency when viewing the identical spatial location at different timesteps. Regarding evaluation metrics, we employ the Fréchet Video Distance (FVD)~\cite{unterthiner2018towards} to assess video quality. Additionally, we utilize PSNR, SSIM~\cite{wang2004image}, and LPIPS~\cite{zhang2018unreasonable} to evaluate both image quality and multi-view consistency. Furthermore, we incorporate the Reprojection Error (RPE) metric~\cite{wu2025geometry} to evaluate geometry-aware consistency. We consider two variants: intra-agent RPE and inter-agent RPE. Intra-agent RPE is computed between overlapping frame pairs within the same generated video, measuring whether a single agent preserves stable scene geometry during continuous observation or revisits. Inter-agent RPE is computed between frame pairs from different agents using their camera poses, measuring whether generated videos maintain consistent scene content in shared spatial regions. 

\noindent\textbf{Baselines.}
We compare our proposed method against two baseline frameworks: (1) State-of-the-Art Single-Agent World Models (Lingbot-World~\cite{lingbot-world}): We directly extend the current single-agent world model to facilitate multi-agent video generation. (2) Concurrent Multi-Agent Models (MultiWorld~\cite{wu2026multiworld}): We compare our approach against concurrent work that similarly tackle the task of multi-agent video generation. 

\begin{table*}[t]
    \centering
    % \small
    % \setlength{\tabcolsep}{2.5pt}
    \caption{\textbf{Quantitative Comparison Results.}~Although the inherent variations between the generated content and the GT result in relatively low absolute scores for pixel-level metrics like PSNR, our method still achieves the best performance across video quality and geometry-aware consistency metrics, demonstrating stronger camera control and multi-view consistency than existing baselines.}
    \begin{tabular}{lcccccc}
        \toprule
        Method & PSNR$\uparrow$ & SSIM$\uparrow$ & LPIPS$\downarrow$ & FVD$\downarrow$ & Intra-RPE$\downarrow$ & Inter-RPE$\downarrow$ \\
        \midrule
        Lingbot~\cite{lingbot-world} & 10.47 & 0.377 & 0.599 & 560.49 & 0.068 & 0.219 \\
        MultiWorld~\cite{wu2026multiworld} & 10.64 & 0.366 & 0.579 & 782.34 & 0.074 & 0.190 \\
        Ours & \textbf{14.12} & \textbf{0.488} & \textbf{0.360} & \textbf{134.9} & \textbf{0.022} & \textbf{0.084} \\
        \bottomrule
    \end{tabular}
    \label{tab:quantitative_compare}
\end{table*}

\begin{figure*}[t]
    \centering
    \includegraphics[width=\linewidth]{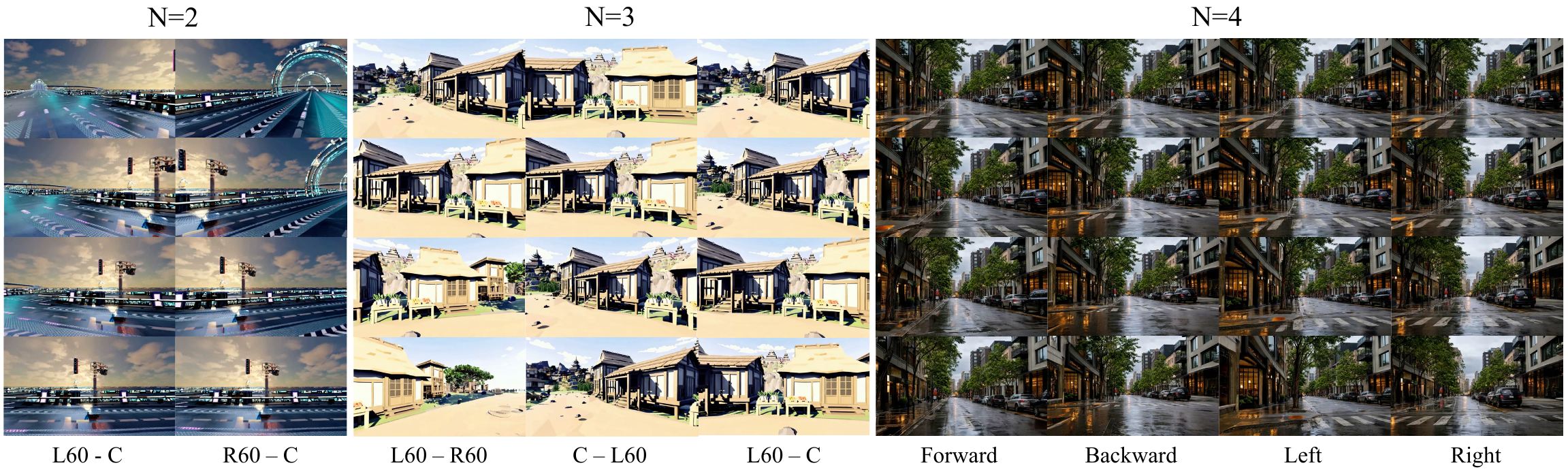}
    \caption{
        \textbf{Scalability of agent number.}~Prisma-World allows for the arbitrary specification of the number of output agents across diverse and complex scenes, while robustly preserving both the visual quality and multi-view consistency in output videos.
    } 
    \label{fig:scalable}
    \vspace{-3mm}
\end{figure*}

\subsection{Quantitative Results}
Table~\ref{tab:quantitative_compare} presents the quantitative comparison between our approach and the baseline methods. As observed, single-agent models inherently lack dedicated mechanisms for multi-agent processing, thereby failing to preserve multi-agent view consistency in their generated results. Furthermore, while existing multi-agent approaches attempt to address this, they are predominantly constrained to specific, isolated game environments, rendering them incapable of yielding satisfactory outcomes when generalizing to more complex and diverse scenes. In contrast, our proposed method consistently generates high-quality multi-agent videos across highly complex environments while robustly enforcing strict multi-view consistency. 

\subsection{Qualitative Results}
A qualitative comparison between our proposed method and the baseline frameworks is illustrated in Fig.~\ref{fig:compare}. As demonstrated, our approach successfully synthesizes high-quality multi-agent videos while strictly preserving multi-view consistency. Furthermore, Fig.~\ref{fig:scalable} highlights our model's capacity to flexibly specify the number of agents at inference time. Utilizing a single pre-trained model, we can generate high-quality multi-agent outputs across diverse agent numbers while robustly maintaining multi-view consistency among all participating agents.

\begin{table*}[t]
    \centering
    % \small
    % \setlength{\tabcolsep}{2.5pt}
    \caption{\textbf{Ablation on RoPE and overlap-decaying curriculum training.}~}
    \begin{tabular}{lcccccc}
        \toprule
        Method & PSNR$\uparrow$ & SSIM$\uparrow$ & LPIPS$\downarrow$ & FVD$\downarrow$ & Intra-RPE$\downarrow$ & Inter-RPE$\downarrow$ \\
        \midrule
        w/o MA-RoPE & 13.58 & 0.468 & 0.372 & 171.23 & 0.024 & 0.113 \\
        w/o odc & 13.67 & 0.471 & 0.368 & 166.17 & 0.025 & 0.108\\
        Ours & \textbf{14.12} & \textbf{0.488} & \textbf{0.360} & \textbf{134.98} & \textbf{0.022} & \textbf{0.084} \\
        \bottomrule
    \end{tabular}
    \label{tab:abla}
\end{table*}

\subsection{Ablation Study}
\begin{figure*}[t]
    \centering
    \includegraphics[width=\linewidth]{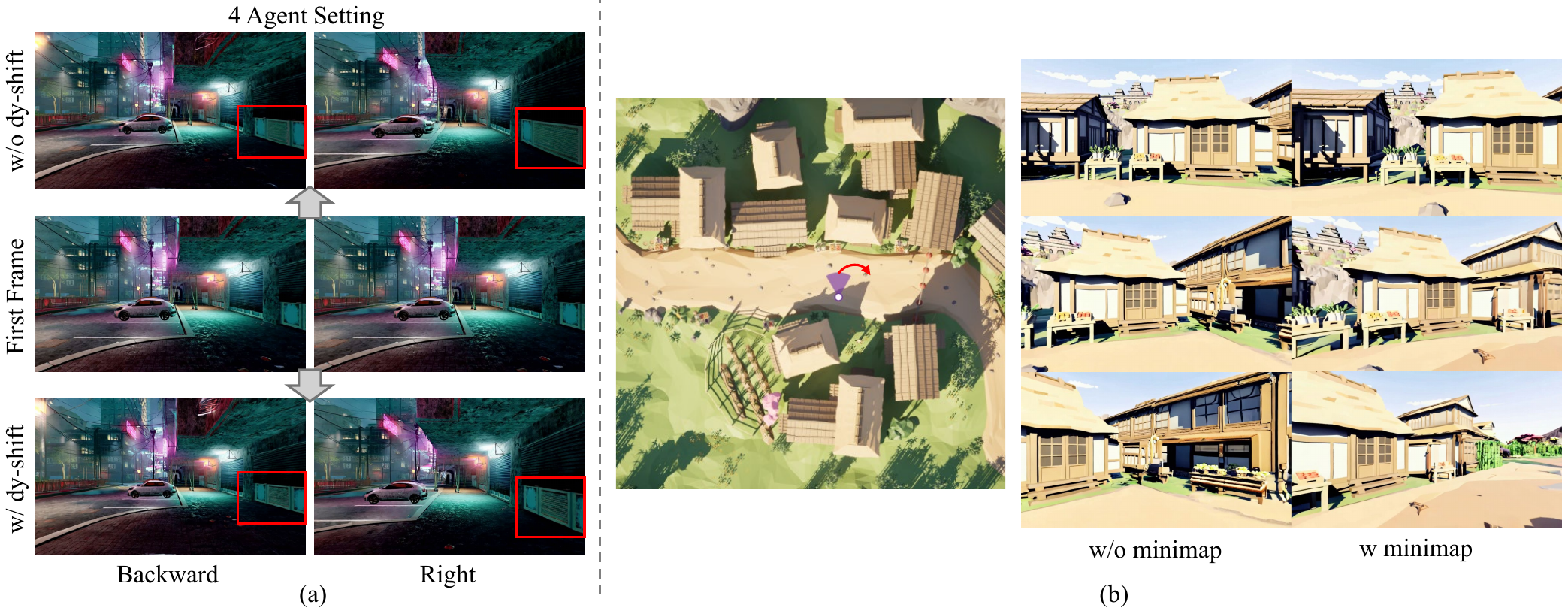}
    \caption{
        \textbf{Ablation on dynamic shift and minimap.}~
    } 
    \label{fig:ablation}
\end{figure*}

To evaluate the effectiveness of the individual components and training strategies within our proposed framework, we conducted the following ablation studies: (1) \textbf{RoPE Design}: We substitute our multi-agent RoPE with the standard $3$D RoPE, essentially processing the concatenated multi-agent video sequence as a single, continuous video (w/o MA-RoPE). (2) \textbf{Overlap-Decaying Curriculum Training}: We discard the progressive curriculum learning strategy, initializing the training phase directly and exclusively with hard samples (w/o odc). (3) \textbf{Dynamic Noise Shift}: We enforce a static noise shift parameter during training, completely disregarding the fluctuating number of agents within the batch (w/o dy-shift). (4) \textbf{Minimap}: We compare the results before and after introducing the minimap condition. The quantitative results are detailed in Table~\ref{tab:abla}. It is evident that removing either the RoPE design or the curriculum strategy consistently degrades video quality and consistency, confirming their contribution to joint multi-agent generation. Furthermore, as illustrated in Fig.~\ref{fig:ablation}(a), maintaining a static noise shift leads to a degradation in multi-view consistency as the number of agents increases. In contrast, the implementation of a dynamic noise shift schedule robustly ensure high-quality and consistency generation across variable agent counts. To demonstrate the function of the minimap input, we also conducted an ablation study on this conditioning signal. As evidenced by the results in Fig.~\ref{fig:ablation}(b), the minimap effectively guides the video generation process by providing the model with explicit scene structural information regarding the immediate surroundings of the agent. As illustrated in the figure, an agent is positioned on a horizontal yellow path and is oriented upwards. We condition the model to generate a video sequence in which the agent rotates to the right. In the absence of minimap guidance, the model hallucinates novel scene content based solely on its inherent generative capabilities. Conversely, upon incorporating the local information from the minimap, the model correctly infers the presence of a yellow path to the right. Consequently, it faithfully reflects this environmental feature within the synthesized video, successfully demonstrating the effective guidance of local spatial structural information.

\section{Conclusion}
This paper introduces \textbf{Prisma-World}, a camera-controllable multi-agent world model that generates multiple agents as consistent observations of a shared scene. Instead of sampling each video independently, Prisma-World performs joint geometry-aware denoising over all agent videos, using full attention, MA-RoPE, and relative camera geometry to couple overlapping viewpoints during generation. We further incorporate minimap-conditioned spatial guidance and construct PrismaDataset with panoramic acquisition, flexible multi-agent view composition, and precise camera/action annotations. Experiments show that Prisma-World generates high-quality videos with variable agent numbers while improving multi-view consistency. Future work will extend this framework toward longer-horizon interaction, richer dynamic objects, and real-world multi-agent data.

{
    \small
    \bibliographystyle{unsrtnat}
    \bibliography{neurips_2026}
}

% {
% \small

% [1] Alexander, J.A.\ \& Mozer, M.C.\ (1995) Template-based algorithms for
% connectionist rule extraction. In G.\ Tesauro, D.S.\ Touretzky and T.K.\ Leen
% (eds.), {\it Advances in Neural Information Processing Systems 7},
% pp.\ 609--616. Cambridge, MA: MIT Press.

% [2] Bower, J.M.\ \& Beeman, D.\ (1995) {\it The Book of GENESIS: Exploring
%   Realistic Neural Models with the GEneral NEural SImulation System.}  New York:
% TELOS/Springer--Verlag.

% [3] Hasselmo, M.E., Schnell, E.\ \& Barkai, E.\ (1995) Dynamics of learning and
% recall at excitatory recurrent synapses and cholinergic modulation in rat
% hippocampal region CA3. {\it Journal of Neuroscience} {\bf 15}(7):5249-5262.
% }

%%%%%%%%%%%%%%%%%%%%%%%%%%%%%%%%%%%%%%%%%%%%%%%%%%%%%%%%%%%%

% \appendix

% \section{Technical appendices and supplementary material}
% Technical appendices with additional results, figures, graphs, and proofs may be submitted with the paper submission before the full submission deadline (see above). You can upload a ZIP file for videos or code, but do not upload a separate PDF file for the appendix. There is no page limit for the technical appendices. 

% Note: Think of the appendix as ``optional reading'' for reviewers. The paper must be able to stand alone without the appendix; for example, adding critical experiments that support the main claims to an appendix is inappropriate. 

%%%%%%%%%%%%%%%%%%%%%%%%%%%%%%%%%%%%%%%%%%%%%%%%%%%%%%%%%%%%

% \newpage
% \input{checklist.tex}

\end{document}